\documentclass[conference]{IEEEtran}
\IEEEoverridecommandlockouts
\usepackage{cite}
\usepackage{amsmath,amssymb,amsfonts}
\usepackage{algorithmic}
\usepackage{graphicx}
\usepackage{hyperref}
\usepackage{textcomp}
\usepackage{xcolor}
\usepackage[top=0.75in, bottom=1.09in, left=0.65in, right=0.65in, columnsep=0.24in]{geometry}
\def\BibTeX{{\rm B\kern-.05em{\sc i\kern-.025em b}\kern-.08em
    T\kern-.1667em\lower.7ex\hbox{E}\kern-.125emX}}
\begin{document}

\title{Comparing Reconstruction Attacks on Pretrained Versus Full Fine-tuned Large Language Model Embeddings on Homo Sapiens Splice Sites Genomic Data
}

\author{\IEEEauthorblockN{1\textsuperscript{st} Reem Al-Saidi}
\IEEEauthorblockA{\textit{School of Computer Science} \\
\textit{University of Windsor}\\
Windsor, Canada \\
alsaidir@uwindsor.ca}
\and
\IEEEauthorblockN{2\textsuperscript{nd} Erman Ayday}
\IEEEauthorblockA{\textit{Department of Computer and Data Sciences
} \\
\textit{Case Western Reserve University}\\
Cleveland, OH, USA \\
exa208@case.edu}
\and 
\IEEEauthorblockN{3\textsuperscript{rd} Ziad Kobti}
\IEEEauthorblockA{\textit{School of Computer Science} \\
\textit{University of Windsor}\\
Windsor, Canada \\
kobti@uwindsor.ca}
}

\maketitle

\begin{abstract}
This study investigates embedding reconstruction attacks in large language models (LLMs) applied to genomic sequences, with a specific focus on how fine-tuning affects vulnerability to these attacks. Building upon Pan et al.'s seminal work demonstrating that embeddings from pretrained language models can leak sensitive information, we conduct a comprehensive analysis using the HS3D genomic dataset to determine whether task-specific optimization strengthens or weakens privacy protections.
Our research extends Pan et al.'s work in three significant dimensions. 
First, we apply their reconstruction attack pipeline to pretrained and fine-tuned model embeddings, addressing a critical gap in their methodology that did not specify embedding types. Second, we implement specialized tokenization mechanisms tailored specifically for DNA sequences, enhancing the model's ability to process genomic data, as these models are pretrained on natural language and not DNA. Third, we perform a detailed comparative analysis examining position-specific, nucleotide-type, and privacy changes between pretrained and fine-tuned embeddings. We assess embeddings vulnerabilities across different types and dimensions, providing deeper insights into how task adaptation shifts privacy risks throughout genomic sequences.
Our findings show a clear distinction in reconstruction vulnerability between pretrained and fine-tuned embeddings. Notably, fine-tuning strengthens resistance to reconstruction attacks in multiple architectures---XLNet (+19.8\%), GPT-2 (+9.8\%), and BERT (+7.8\%)---pointing to task-specific optimization as a potential privacy enhancement mechanism. These results highlight the need for advanced protective mechanisms for language models processing sensitive genomic data, while highlighting fine-tuning as a potential privacy-enhancing technique worth further exploration.
\end{abstract}


\section{Introduction}

Recent advances in large language models (LLMs) have revolutionized computational genomics \cite{ali2025large}, enabling sophisticated analysis of DNA sequences through vector representations known as embeddings \cite{sarumi2024large}. These embeddings power critical applications, including splice site prediction, gene expression modeling, and genetic variant classification \cite{minai2025context}, \cite{avsec2021effective}. However, as these powerful representation processes increasingly sensitive genomic data—information that reveals health predispositions, ancestry, and individual identity—questions of privacy protection become paramount. The potential privacy risks inherent in model embeddings represent a significant yet underexplored concern for genome applications.

Genomic data privacy is especially important because genetic information has a uniquely sensitive and immutable nature \cite{naveed2015privacy}. DNA sequences can show a person's risk for certain diseases, their ancestry information, and even personality traits \cite{naveed2015privacy}, \cite{sariyar2017sensitive}. Pan et al \cite{pan2020privacy} demonstrated a concerning vulnerability: \textit{embeddings from general-purpose language models can be reverse-engineered to reconstruct the original input data with substantial accuracy}. Their research showed that attackers could recover genomic sequences from the Homo Sapiens Splice Sites Dataset (HS3D) \cite{pollastro2002hs3d} simply by using embeddings produced by pre-trained models such as BERT, RoBERTa, XLNet, and GPT-2 \cite{pan2020privacy}. These findings raised critical privacy concerns for models handling sensitive genetic information, particularly due to their increasing adoption in healthcare settings where data protection is essential \cite{casey2025privacy}.
Prior research on embedding privacy has focused on text data or image representations, with relatively limited attention to genomic embeddings \cite{song2020information}. Recent work has begun addressing this gap, with Pan et al.'s research highlighting individual-level privacy risks in genomic embeddings \cite{pan2020privacy}, while Al-Saidi et al.~\cite{al2025privacy} investigated privacy vulnerabilities at the population level, examining how embeddings might leak sensitive demographic or population-specific genetic information.
A significant knowledge gap exists regarding how model specialization affects privacy vulnerabilities. While Pan et al.'s research examined general-purpose LLMs used as feature extractors without domain adaptation \cite{pan2020privacy}, the effects of fine-tuning on privacy leakage remain unexplored. This gap is significant because fine-tuning is the standard approach when applying language models to specialized domains like genomics. Understanding whether embeddings extracted from fine-tuned models mitigate or exacerbate privacy risks is crucial for developing secure genomic analysis pipelines.
Our research addresses this gap by systematically investigating the domain adaptation of LLMs through fine-tuning on genomic tasks, while customizing the original tokenization schemes to better handle DNA sequences. 

Our research makes several significant contributions to understanding LLM embedding privacy in the computational genomics domain:
\begin{itemize}
\item We provide an investigation of how domain adaptation through fine-tuning affects privacy vulnerabilities in language models processing genomic sequences. Using rigorous empirical analysis, we quantify how specialized genomic fine-tuning embedding persists reconstruction risks compared to pretrained models, revealing the unexpected finding that task-specific optimization enhances privacy protection in most tested models.

\item We address a critical limitation in Pan et al.'s seminal work by explicitly specifying and maintaining consistent embedding types throughout our comparisons. While Pan et al. used the model's original tokenization and did not specify which embedding representations were used in their genome reconstruction attacks: 
\begin{itemize}
    \item Our study rigorously controls for embedding type when comparing pretrained and fine-tuned models, ensuring valid and reproducible privacy vulnerability assessments.
    \item We customize the models' tokenization mechanisms to explicitly handle DNA sequences, which enhances the models' ability to process genomic information while allowing for meaningful privacy analysis.
\end{itemize}

\item We significantly expand the evaluation metrics beyond Pan et al.'s original metrics by implementing: (1) Nucleotide-specific reconstruction, (2) Privacy changes between the pretrained and fine-tuned embeddings. 
The extended metrics enable direct assessment of how domain adaptation impacts privacy risks across different aspects of genomic data.

\end{itemize}

These contributions enhance understanding of the privacy implications of LLMs in genomic applications by providing theoretical consideration and practical direction for researchers dealing with sensitive genetic information.
The remainder of this paper is organized as follows: Section II reviews related work in embedding privacy and genomic data protection. Section III provides an overview of language model architectures. Section IV shows the original model's tokenization mechanism and our customized one. Moreover, it highlights different embedding types examined in our privacy analysis. Section V details our experimental methodology, including the dataset, fine-tuning procedures, and evaluation metrics. Section VI presents our findings and analysis through a detailed discussion of the results, while Section VII concludes with future research directions.
\section{Related Work}

Language models can unintentionally memorize and leak sensitive training data~\cite{satvaty2024undesirable}, where reconstruction attacks can be performed via model query. This vulnerability extends to embeddings, which can reveal substantial information about the original input text~\cite{song2020information}, \cite{wan2024information}. Pan et al. \cite{pan2020privacy} pioneered in the reconstruction attacks across various LLM embeddings by showing how genomic sequence data could be reconstructed from its embeddings.
Al-Saidi et al.~\cite{al2025privacy} assessed DNABert-S embeddings for reconstruction attacks at the population level, extending privacy concerns beyond individual-level risks. Despite established privacy risks in language model embeddings across different modalities and applications \cite{team2024unveiling}, a critical knowledge gap exists regarding how fine-tuning affects these vulnerabilities in genomic applications. Our work aims to address the privacy of fine-tuning embeddings.

\section{Background}
\subsection{Pretrained Model Architectures}

The models in our study span several architectural categories: bidirectional encoders, autoregressive decoders, and models with specialized pretraining objectives.

BERT~\cite{devlin2019bert}, the foundational bidirectional encoder, employs WordPiece tokenization with a 30,522 token vocabulary. Its architecture consists of 12 transformer layers, 768-dimensional hidden representations, and 12 attention heads in the base version. 

XLNet~\cite{yang2019xlnet} introduces a permutation-based pretraining approach with SentencePiece tokenization. Its base configuration mirrors BERT with 12 layers and 768-dimensional embeddings, but processes information differently through its autoregressive formulation. 

The GPT-2~\cite{radford2019language} is an autoregressive decoder-only model. It uses byte-level BPE with 50,257 tokens. Both models feature 12 layers and 768-dimensional representations.

RoBERTa~\cite{liu2019roberta}, a robustly optimized BERT variant, uses byte-level BPE tokenization with 50,265 tokens. Its architecture matches BERT-base with 12 layers and 768-dimensional representations.

ALBERT~\cite{lan2019albert} employs a parameter-efficient design with factorized embedding parameterization, projecting from a 128-dimensional embedding space to the 768-dimensional hidden space. Using SentencePiece tokenization with 30,000 tokens.

ERNIE~\cite{zhang2019ernie} enhances BERT's architecture with knowledge integration, using WordPiece tokenization with 30,522 tokens. The model maintains BERT's dimensions (768) and layer count (12).

\section{Tokenization and Embedding Process for Genomic Sequences}

For processing genomic sequences, we modify language model tokenizers—created for natural language—to work efficiently with DNA's limited four nucleotides:  (A),  (C),  (G), and  (T). This adjustment allows for proper tokenization of genetic data, even though the structure of DNA is quite different from regular text.

\subsection{Tokenization Methods and Customization}
For a genomic DNA sequence $S = (s_1, s_2, \ldots, s_n)$ where $s_i \in \{A, C, G, T\}$, each model's tokenizer $\tau$ processes the sequence into tokens:

\begin{equation}
\tau: S \mapsto T, \quad \text{where} \quad T = \tau(S) = (t_1, t_2, \ldots, t_m)
\end{equation}
Below, we describe the original tokenization methods for each model type and our standardized customization approach.
\subsubsection{WordPiece Tokenization (WP)}

\textbf{\\Original Method:} In BERT\cite{devlin2019bert} and ERNIE\cite{zhang2019ernie}, WP splits text into subword units through a greedy longest-match-first approach\cite{devlin2019bert,zhang2019ernie}. It maximizes the likelihood of the training data given the final vocabulary:

\begin{equation}
\tau_{\text{WP}}(S) = \underset{(t_1,\ldots,t_m)}{\arg\max} \prod_{i=1}^{m} P(t_i|t_1,\ldots,t_{i-1})
\end{equation}

When applied to DNA sequences, WordPiece would likely process them into meaningless subword units like "AC", "GT", or "CG", which disrupts the biological meaning. For example, a DNA sequence "ACGTAACGT" might be tokenized as:

\begin{equation}
\tau_{\text{WP}}(\text{ACGTAACGT}) = (\text{AC}, \text{CG}, \text{GTA}, \text{TA})
\end{equation}
\textbf{\\Customized Approach} To maintain the biological feature and consistent tokenization, we modified WordPiece tokenizers as follows:
\begin{itemize}
    \item Extended vocabulary with nucleotide-specific tokens:
    \begin{equation}
    \mathcal{V}_{\text{WP}}^{\text{new}} = \mathcal{V}_{\text{WP}}^{\text{orig}} \cup \{A, C, G, T\}
    \end{equation}
    \item Space-separated preprocessing to force tokenization at the nucleotide level:
    \begin{equation}
    S_{\text{processed}} = join(S, \text{ ``~''}) = s_1 \text{ } s_2 \text{ } \ldots \text{ } s_n
    \end{equation}
 For example, a DNA sequence "ACGTAACGT" is transformed to "A C G T A A C G T", ensuring each nucleotide is processed as a separate token rather than being merged into biologically meaningless subwords.
    \item Resized embedding matrix to accommodate the new vocabulary:
    \begin{equation}
    E_{\text{new}} \in \mathbb{R}^{|\mathcal{V}_{\text{WP}}^{\text{new}}| \times d}
    \end{equation}
\end{itemize}

This results in consistent nucleotide-level tokenization:
\begin{equation}
\tau_{\text{WP}}^{\text{custom}}(\text{ACGTAACGT}) = (A, C, G, T, A, A, C, G, T)
\end{equation}
\subsubsection{SentencePiece Tokenization (SP)}
\textbf{\\Original Method:} In XLNet\cite{yang2019xlnet} and ALBERT\cite{lan2019albert}, SP applies an unigram language model to determine the most likely segmentation\cite{yang2019xlnet,lan2019albert}:
\begin{equation}
\tau_{\text{SP}}(S) = \underset{T}{\arg\max} P(T|S) = \underset{T}{\arg\max} \prod_{i=1}^{m} P(t_i)
\end{equation}
For DNA sequences, SentencePiece might process them as unknown tokens, falling back to character-level encoding or using special unknown token symbols. For example:
\begin{equation}
\tau_{\scriptsize{\text{SP}}}(\text{ACGT}) = 
\begin{cases}
(\text{UNK}_1, \ldots) \text{\scriptsize{ (unknown)}} \\
(\text{A}, \text{CG}, \ldots) \text{\scriptsize{ (arbitrary)}}
\end{cases}
\end{equation}

This unpredictable tokenization could create biological information loss or arbitrary subword units that don't align with nucleotide boundaries.

\textbf{\\Customized Approach} For SentencePiece-based models like XLNet\cite{yang2019xlnet} and ALBERT\cite{lan2019albert}, we applied a similar customization strategy:
\begin{itemize}
    \item Extended vocabulary with nucleotide tokens:
    \begin{equation}
    \mathcal{V}_{\text{SP}}^{\text{new}} = \mathcal{V}_{\text{SP}}^{\text{orig}} \cup \{A, C, G, T\}
    \end{equation}
    \item Space-separated preprocessing:
    \begin{equation}
    S_{\text{processed}} = join(S, \text{ ``~''}) = s_1 \text{ } s_2 \text{ } \ldots \text{ } s_n
    \end{equation}
    \item Consistent tokenization parameters:
    \begin{equation}
    \begin{aligned}
    \text{max\_length} &= 60 \\
    \text{padding} &= \text{'max\_length'} \\
    \text{truncation} &= \text{True}
    \end{aligned}
    \end{equation}
\end{itemize}
For ALBERT, this customization maintains the model's parameter-sharing architecture while ensuring nucleotide-level granularity:
\begin{equation}
\tau_{\text{ALBERT}}^{\text{custom}}(\text{ACGTAA}) = (A, C, G, T, A, A)
\end{equation}
\subsubsection{Byte Per Encoding (BPE) }

\textbf{\\Original Method:} In GPT-2\cite{radford2019language}, and RoBERTa\cite{liu2019roberta}), BPE iteratively merges the most frequent character or subword pairs\cite{radford2018improving,radford2019language,liu2019roberta}:

\begin{equation}
\tau_{\text{BPE}}(S) = \text{Apply}(\{(a,b) \rightarrow ab\}, S)
\end{equation}

where the pairs $(a,b)$ are selected according to frequency statistics from the training data.

For DNA sequences, BPE might recognize short n-grams of nucleotides that happen to match existing tokens in their vocabulary. For example:

\begin{equation}
\tau_{\text{BPE}}(\text{ACGTAACGT}) = (\text{AC}, \text{G}, \text{TA}, \text{AC}, \text{G}, \text{T})
\end{equation}

If certain pairs like "AC" or "TA" were frequent in the training data. This results in inconsistent segmentation that doesn't respect biological boundaries.

\textbf{\\Customized Approach} For BPE-based models, our customization includes:

\begin{itemize}
    \item Extended vocabulary with nucleotide tokens:
    \begin{equation}
    \mathcal{V}_{\text{BPE}}^{\text{new}} = \mathcal{V}_{\text{BPE}}^{\text{orig}} \cup \{A, C, G, T\}
    \end{equation}
    \item Space-separated preprocessing:
    \begin{equation}
    S_{\text{processed}} = join(S, \text{ ``~''}) = s_1 \text{ } s_2 \text{ } \ldots \text{ } s_n
    \end{equation}
    \item Padding token adaptation for GPT-2 model:
    \begin{equation}
    \text{pad\_token} = eos\_token \quad \text{(for GPT-2 only)}
    \end{equation}
\end{itemize}
This customization ensures consistent tokenization across all BPE-based models:
\begin{equation}
\tau_{\text{BPE}}^{\text{custom}}(\text{ACGTAACGT}) = (A, C, G, T, A, A, C, G, T)
\end{equation}
\subsection{Token Encoding and Embedding Process}
After applying our customized tokenization to genomic sequences, the tokens are processed through the standard embedding and encoding pipeline of transformer-based language models:

\subsubsection{Token Encoding}
After tokenization, each token $t_i$ is converted to its corresponding one-hot vector $\mathbf{v}_i$:
\begin{equation}
\mathbf{v}_i = \text{OneHot}(t_i) \in \{0,1\}^{|\mathcal{V}|}
\end{equation}
where $\mathcal{V}$ is the vocabulary of the model.
\begin{itemize}
    \item Initial Embedding:
The one-hot vectors are transformed into dense embedding vectors by multiplication with the token embedding matrix $E \in \mathbb{R}^{|\mathcal{V}| \times d}$:
\begin{equation}
\mathbf{e}_i^{(0)} = E \cdot \mathbf{v}_i + \mathbf{p}_i
\end{equation}
where $\mathbf{e}_i^{(0)}$ is the initial embedding vector for token $i$, $\mathbf{p}_i$ is the positional encoding for position $i$, and $d$ is the embedding dimension.

\item Contextual Embedding:
These initial embeddings form a sequence $\mathbf{E}^{(0)} = [\mathbf{e}_1^{(0)}, \mathbf{e}_2^{(0)}, \ldots, \mathbf{e}_m^{(0)}]$ that is processed through the model's $L$ transformer layers:
\begin{equation}
\mathbf{E}^{(l)} = f_l(\mathbf{E}^{(l-1)}), \quad \text{for} \quad l = 1, 2, \ldots, L
\end{equation}
where $m$ is the sequence length after tokenization and $\mathbf{E}^{(l)}$ represents the sequence of token embeddings after layer $l$.
\item Final Representation:
The final embedding representation used for downstream tasks is extracted from $\mathbf{E}^{(L)}$ using one of the following methods:
\begin{itemize}
    \item CLS Token Embedding (BERT~\cite{devlin2019bert}, RoBERTa~\cite{liu2019roberta}, ALBERT~\cite{lan2019albert}, ERNIE~\cite{zhang2019ernie}):
\begin{equation}
\mathbf{e}_{\text{cls}} = \mathbf{e}^{(L)}_{\text{[CLS]}}
\end{equation}
where $\mathbf{e}^{(L)}_{\text{[CLS]}}$ is the final representation of the special classification token.

\item Last Token (GPT-2~\cite{radford2019language}, XLNet~\cite{yang2019xlnet}):
\begin{equation}
\mathbf{e}_{\text{last}} = \mathbf{e}^{(L)}_m
\end{equation}
where $\mathbf{e}^{(L)}_m$ is the final representation of the last token in the sequence.
\end{itemize}
\end{itemize}
\subsection{Fine-tuning Approaches}

Fine-tuning adapts pretrained language models to specific downstream tasks \cite{machacek2025impact}. In full fine-tuning, all parameters of the pretrained model are updated during training on the target task \cite{machacek2025impact}. For a model with parameters $\theta$, the fine-tuning process updates these parameters to $\theta'$ by minimizing a task-specific loss function $\mathcal{L}$:

\begin{equation}
\theta' = \arg\min_{\theta} \mathcal{L}(\theta, \mathcal{D}_{\text{task}})
\end{equation}

where $\mathcal{D}_{\text{task}}$ is the task-specific dataset (in our case, HS3D splice site prediction).
\section{Experimental Design}

\subsection{Dataset Description}

We utilized the Homo Sapiens Splice Sites Dataset (HS3D) \cite{pollastro2002hs3d}, a standard benchmark for splice site prediction, following Pan et al.'s~\cite{pan2020privacy} methodology by extracting 20-nucleotide windows centered on potential splice sites (positions 60-80) from each original sequence. After quality filtering to retain only sequences with exactly 20 canonical nucleotides (A, C, G, T) and no ambiguous bases, we created a training set of 31,680 samples (2,880 positive true splice sites, 28,800 negative false splice sites) and a balanced testing set of 2,000 samples (1,000 positive, 1,000 negative). 

\subsection{Fine-tuning parameters} 
All eight transformer models underwent full parameter fine-tuning for DNA splice site prediction using consistent core hyperparameters: 3 training epochs, 0.01 weight decay, and the Adam optimizer. Learning rates were model-specific, ranging from 1e-5 (BERT-large, ALBERT, RoBERTa) to 5e-5 (GPT2), with batch sizes varying from 16 to 32 and gradient accumulation implemented to maintain effective training batch sizes. DNA-specific preprocessing included adding nucleotide tokens (A, C, G, T) to each vocabulary and spacing nucleotides for individual tokenization, with a maximum sequence length of 60. Model embeddings were extracted from architecture-appropriate positions (CLS token for bidirectional models, first/last token for unidirectional models) and fed into consistent 3-layer MLP classifiers with sigmoid activations.
Table~\ref{tab:hyperparams} highlights the key differentiating hyperparameters across models.
\begin{table}[t]
\renewcommand{\arraystretch}{1.1}
\caption{Model Fine-tuning Configurations}
\label{tab:hyperparams}
\centering
\setlength{\tabcolsep}{4pt}
\begin{tabular}{|l|c|c|c|p{2.8cm}|}
\hline
\textbf{Model} & \textbf{LR} & \textbf{BS/GA} & \textbf{Size} & \textbf{Key Features} \\
\hline
BERT & 3e-5 & 32/- & 110M & Bidirectional \\
\hline
ALBERT & 1e-5 & 32/2 & 12M & Param sharing \\
\hline
ERNIE & 2e-5 & 32/2 & 110M & Knowledge integ. \\
\hline
RoBERTa & 1e-5 & 32/2 & 125M & Opt. masking\\
\hline
GPT2 & 5e-5 & 24/- & 124M & Improved GPT \\
\hline
XLNet & 2e-5 & 16/- & 110M & Permutation LM \\
\hline
\end{tabular}

\vspace{0.1cm}
\footnotesize LR = Learning Rate, BS = Batch Size, GA = Gradient Accumulation steps,\\
\footnotesize Opt = Optimized, LM= Language Modeling, Parm = Parameter.
\end{table}
\subsection{Embedding Analysis and Attack Training}
\subsubsection{Embedding Extraction and Dimensions}
We extract embeddings from both pretrained and fine-tuned models as follows: 

For bidirectional models (BERT, RoBERTa, ALBERT, ERNIE), we extract the first token ([CLS]) representation from the final layer, which encodes sequence-level information. For autoregressive models (GPT-2) and XLNet, we extract the last token representation, which contains information about the entire sequence processed left-to-right. All embeddings are taken from the final layer of each model to capture the most refined representations.

The dimensionality of these embeddings corresponds to each model's hidden dimension: 768 for most models, with XL and XLM producing 1024-dimensional vectors.
\subsection{Attack Methodology and Training}
Our implementation of Pan et al.'s reconstruction attack \cite{pan2020privacy} comprises the same systematic pipeline that trains position-specific classifiers to recover nucleotides at specific positions from embeddings. For each nucleotide position $(1-20)$, we first generate sinusoidal positional embeddings (following the formula $p_{i,2k} = \sin(i/10000^{2k/d})$, $p_{i,2k+1} = \cos(i/10000^{(2k+1)/d})$) with dimension matching the embedding vectors. These positional embeddings are concatenated with the model's embedding representation to create position-aware inputs ($\mathbf{z} \oplus \mathbf{p}_i$). A dedicated three-layer MLP classifier ($200$ hidden units per layer, sigmoid activation, batch normalization) is then trained for each position using $80\%$ of the data with $128$-sample mini-batches over $5$ epochs. Each classifier takes the concatenated vector as input and outputs probabilities across four nucleotide classes (A, C, G, T), effectively learning to reconstruct the original sequence one position at a time.  

\subsection{Evaluation Metrics}

\subsubsection{Baseline Metrics from Pan et al. \cite{pan2020privacy}}

\begin{itemize}
    \item \textbf{Position-specific accuracy}: For each position $i$ in the sequence, accuracy is calculated as:
    \begin{equation}
        \text{Acc}_i = \frac{1}{N} \sum_{j=1}^{N} \mathbb{1}(\hat{y}_{i,j} = y_{i,j})
    \end{equation}
    where $\hat{y}_{i,j}$ is the predicted nucleotide at position $i$ for sequence $j$, $y_{i,j}$ is the true nucleotide, and $\mathbb{1}$ is the indicator function. This metric measures reconstruction success at each of the 20 nucleotide positions.
    
    \item \textbf{Random baseline}: All results are compared against random guessing (25\% for the four nucleotides), as established in Pan et al.'s evaluation framework.
\end{itemize}

\subsubsection{Our Extended Evaluation Metric}
We significantly extend Pan et al.'s evaluation methodology with additional metrics to enable more comprehensive privacy analysis for both pretrained and fine-tuned embeddings:

\begin{itemize}
    \item \textbf{Nucleotide-specific accuracy}: We calculate reconstruction accuracy for each nucleotide type:
    \begin{equation}
        \text{Acc}_{\text{nuc}} = \frac{1}{N_{\text{nuc}}} \sum_{i=1}^{20} \sum_{j=1}^{N} \mathbb{1}(\hat{y}_{i,j} = y_{i,j}) \cdot \mathbb{1}(y_{i,j} = \text{nuc})
    \end{equation}
    where nuc $\in$ \{A, C, G, T\} and $N_{\text{nuc}}$ is the total count of that nucleotide. This reveals whether certain nucleotides leak more information than others.

    \item \textbf{Privacy change measurement}: We quantify both position-specific and overall privacy impact:
\begin{equation}
\begin{aligned}
\Delta\text{Privacy}_i &= \text{Acc}_{i}^{\text{pretrained}} - \text{Acc}_{i}^{\text{fine-tuned}} \\
\overline{\Delta\text{Privacy}} &= \frac{1}{20} \sum_{i=1}^{20} \Delta\text{Privacy}_i
\end{aligned}
\end{equation}

Positive values indicate improved privacy after fine-tuning (decreased reconstruction accuracy), while negative values indicate privacy degradation. The average across all positions quantifies the overall privacy impact.

\end{itemize}
\section{Discussion, Results, and Takeaways}
\begin{table*}[!t]
\caption{Reconstruction Attack Accuracy Across Models, Positions, and Embedding Types}
\label{tab:reconstruction_accuracy}
\centering
\resizebox{\textwidth}{!}{%
\begin{tabular}{|l|l|c|c|c|c|c|c|c|c|c|c|c|c|c|c|c|c|c|c|c|c|c|}
\hline
\textbf{Model} & \textbf{Embedding} & \textbf{P1} & \textbf{P2} & \textbf{P3} & \textbf{P4} & \textbf{P5} & \textbf{P6} & \textbf{P7} & \textbf{P8} & \textbf{P9} & \textbf{P10} & \textbf{P11} & \textbf{P12} & \textbf{P13} & \textbf{P14} & \textbf{P15} & \textbf{P16} & \textbf{P17} & \textbf{P18} & \textbf{P19} & \textbf{P20} & \textbf{Avg} \\
\hline
BERT-base & Pretrained & 1.000 & .450 & .300 & .320 & .300 & .270 & .280 & .320 & .320 & .250 & .250 & .250 & .280 & .360 & .300 & .300 & .370 & .300 & .740 & 1.000 & .380 \\
\cline{2-23}
 & Finetuned & \textbf{.250} & \textbf{.300} & \textbf{.240} & \textbf{.310} & \textbf{.270} & .310 & \textbf{.270} & \textbf{.260} & \textbf{.260} & .260 & .340 & .340 & .350 & \textbf{.350} & \textbf{.280} & \textbf{.280} & \textbf{.300} & \textbf{.290} & \textbf{.230} & \textbf{.280} & \textbf{.302} \\
\cline{2-23}
 & Pan's & .982 & .910 & .720 & .530 & .490 & .485 & .480 & .475 & .470 & .470 & .470 & .470 & .470 & .485 & .495 & .520 & .570 & .650 & .770 & .920 & .598 \\
\hline
XLNet-Base & Pretrained & .720 & .480 & .360 & .450 & .460 & .440 & .410 & .440 & .420 & .310 & .400 & .400 & .380 & .480 & .400 & .450 & .570 & .400 & .660 & .970 & .471 \\
\cline{2-23}
 & Finetuned & \textbf{.250} & \textbf{.240} & \textbf{.280} & \textbf{.220} & \textbf{.250} & \textbf{.250} & \textbf{.290} & \textbf{.290} & \textbf{.280} & \textbf{.270} & \textbf{.280} & \textbf{.280} & \textbf{.290} & \textbf{.290} & \textbf{.260} & \textbf{.260} & \textbf{.310} & \textbf{.340} & \textbf{.290} & \textbf{.270} & \textbf{.273} \\
\cline{2-23}
 & Pan's & .990 & .930 & .780 & .620 & .520 & .480 & .460 & .450 & .450 & .450 & .450 & .460 & .470 & .490 & .510 & .540 & .590 & .660 & .770 & .900 & .624 \\
\hline
GPT-2 & Pretrained & .820 & .640 & .520 & .250 & .260 & .360 & .350 & .340 & .410 & .280 & .340 & .300 & .360 & .390 & .310 & .300 & .400 & .470 & .960 & 1.000 & .450 \\
\cline{2-23}
 & Finetuned & \textbf{.240} & \textbf{.260} & \textbf{.310} & .250 & .260 & .400 & \textbf{.240} & \textbf{.300} & \textbf{.340} & .290 & \textbf{.270} & .390 & \textbf{.350} & \textbf{.330} & \textbf{.310} & \textbf{.270} & \textbf{.260} & \textbf{.290} & \textbf{.280} & \textbf{.480} & \textbf{.352} \\
\cline{2-23}
 & Pan's & .990 & .940 & .810 & .660 & .550 & .490 & .470 & .460 & .450 & .450 & .460 & .470 & .480 & .500 & .530 & .570 & .620 & .700 & .810 & .950 & .624 \\
\hline
RoBERTa-Base & Pretrained & .250 & .250 & .250 & .250 & .250 & .190 & .340 & .220 & .240 & .240 & .240 & .220 & .250 & .270 & .260 & .260 & .280 & .220 & .450 & .220 & .264 \\
\cline{2-23}
 & Finetuned & .370 & .330 & .280 & .300 & .340 & .300 & \textbf{.260} & .320 & .330 & .340 & .310 & .240 & .350 & .350 & .400 & .280 & .320 & .390 & \textbf{.290} & .470 & .331 \\
\cline{2-23}
 & Pan's & .600 & .510 & .460 & .410 & .390 & .380 & .370 & .360 & .355 & .350 & .350 & .350 & .360 & .370 & .380 & .390 & .410 & .440 & .480 & .550 & .380 \\
\hline
ERNIE-Base & Pretrained & .260 & .250 & .270 & .230 & .250 & .270 & .320 & .230 & .240 & .240 & .290 & .260 & .250 & .230 & .270 & .230 & .240 & .250 & .190 & .250 & .255 \\
\cline{2-23}
 & Finetuned & .280 & .250 & \textbf{.240} & .270 & .250 & .370 & \textbf{.320} & .300 & .300 & .260 & \textbf{.240} & .280 & .270 & .320 & .350 & .270 & .350 & .270 & .270 & .300 & .284 \\
\cline{2-23}
 & Pan's & .760 & .660 & .550 & .430 & .390 & .380 & .370 & .380 & .400 & .430 & .440 & .450 & .460 & .470 & .480 & .490 & .510 & .550 & .580 & .640 & .483 \\
\hline
ALBERT-Base-V2 & Pretrained & .250 & .270 & .280 & .270 & .250 & .280 & .320 & .250 & .250 & .260 & .250 & .220 & .260 & .250 & .270 & .240 & .280 & .290 & .200 & .230 & .258 \\
\cline{2-23}
 & Finetuned & \textbf{.240} & \textbf{.260} & \textbf{.270} & \textbf{.220} & \textbf{.240} & \textbf{.210} & \textbf{.280} & \textbf{.230} & .260 & .280 & \textbf{.200} & .230 & \textbf{.250} & .260 & .270 & .270 & .280 & .310 & .320 & .250 & \textbf{.257} \\
\cline{2-23}
 & Pan's & - & - & - & - & - & - & - & - & - & - & - & - & - & - & - & - & - & - & - & - & - \\
\hline
\end{tabular}%
}
\vspace{0.1cm}
\footnotesize Note: \textbf{Bold values} in fine-tuned rows indicate positions where fine-tuning provides better privacy protection against nucleotide reconstruction.\\
\footnotesize 
\end{table*}

\subsection{ALBERT Model Privacy Analysis}
\begin{figure*}[htbp]
\begin{center}
\begin{tabular}{ccc}
\includegraphics[width=0.30\textwidth]{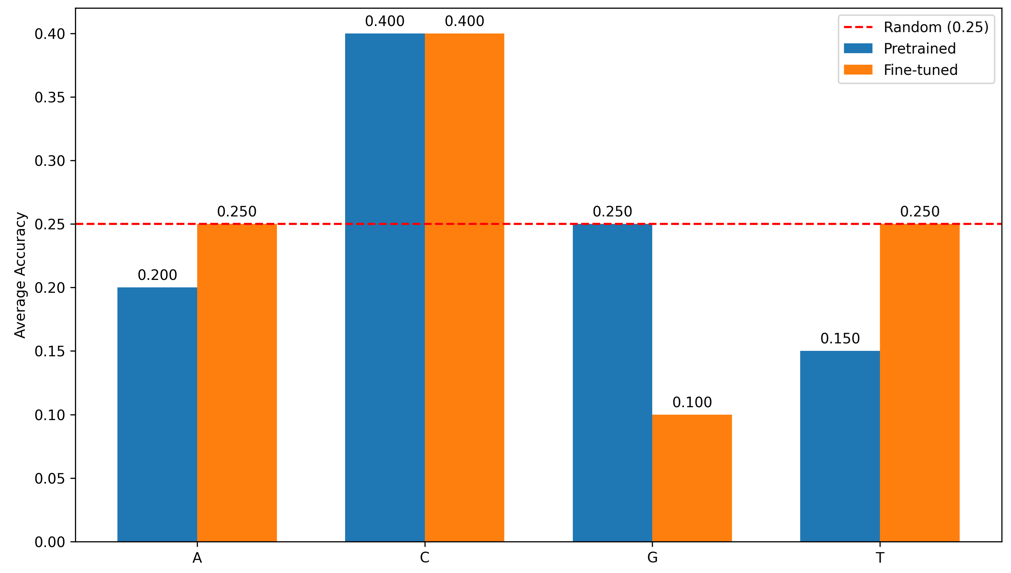} &
\includegraphics[width=0.30\textwidth]{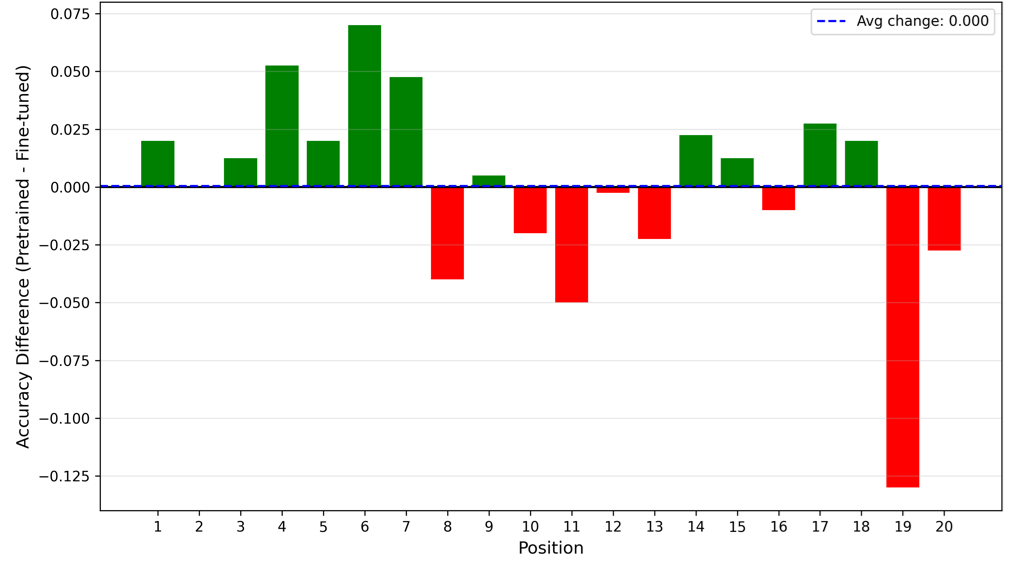} \\
{\fontfamily{ptm}\fontsize{8}{10}\selectfont (a)} & 
{\fontfamily{ptm}\fontsize{8}{10}\selectfont (b)}
\end{tabular}
\end{center}
\caption{Privacy analysis of ALBERT model under nucleotide reconstruction attacks: (a) Nucleotide-wise reconstruction accuracy comparison ; (b) Privacy change after fine-tuning, where positive values (green) indicate improved privacy and negative values (red) indicate decreased privacy}
\label{fig:albert_analysis}
\end{figure*}

The analysis of ALBERT's vulnerability to nucleotide reconstruction attacks reveals a nuanced privacy landscape. As shown in Table~\ref{tab:reconstruction_accuracy}, both pre-trained and fine-tuned models perform only marginally above random chance (0.258 and 0.257, respectively, versus the 0.25 random baseline), indicating limited but present vulnerability. Moreover, the vulnerability varies by position, with neither version showing consistent advantages across all positions.
Fig.~\ref{fig:albert_analysis}(a) breaks down vulnerability by nucleotide type, revealing that  (C) remains equally vulnerable in both versions with 0.4 accuracy (well above the 0.25 random baseline), (G) becomes significantly more private after fine-tuning (dropping from 0.25 to 0.1), while (T) shows decreased privacy (rising from 0.15 to 0.25). For (A) it is become more vulnerable after fine-tuning.

Fig.~\ref{fig:albert_analysis}(b) provides a clearer visualization of these position-specific differences, demonstrating that early sequence positions (1-7) generally become more secure after fine-tuning (green bars) while later positions show mixed results, with position 19 exhibiting the most substantial privacy degradation (red bar reaching approximately -0.125).

\textbf{Key Takeaways for ALBERT}
\begin{enumerate}
    \item Fine-tuning ALBERT creates a zero-sum privacy effect—overall vulnerability remains unchanged while being redistributed across different positions and nucleotides as the ``Avg change: 0.000''.
    
    \item ALBERT demonstrates nucleotide-specific privacy behaviors, with C consistently vulnerable, G becoming more private, T becoming less private after fine-tuning, and A more vulnerable after fine-tuning.
    
    \item Sequence position significantly affects vulnerability, with positions 1-3 and 18-20 generally more susceptible to reconstruction attacks.
    
    \item  The privacy-preserving strategies for ALBERT should focus on position-specific and nucleotide-specific mitigations rather than relying solely on full fine-tuning.
\end{enumerate}

\subsection{BERT Model Privacy Analysis}

\begin{figure*}[!t]
\centering
\begin{tabular}{ccc}

\includegraphics[width=0.30\textwidth]{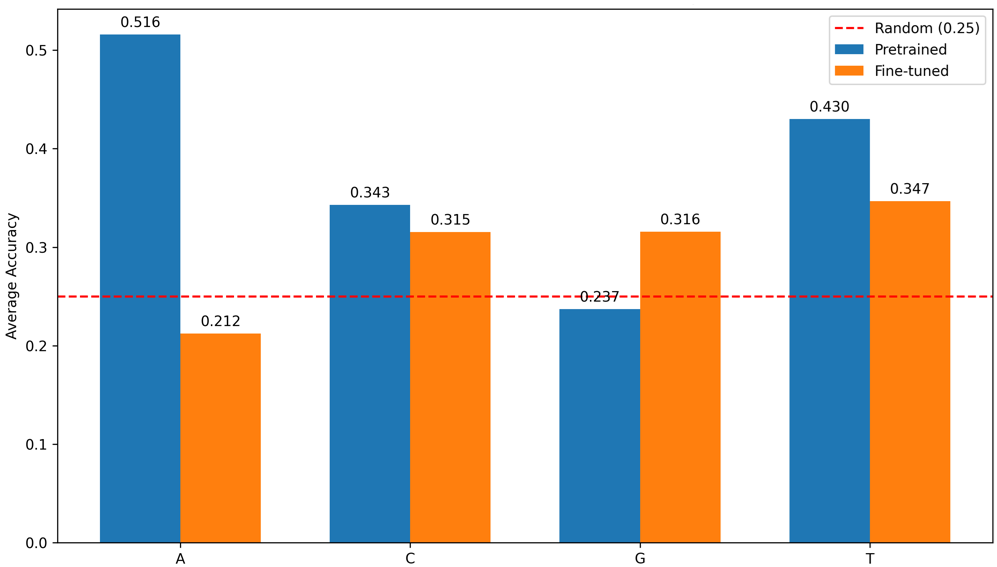} &
\includegraphics[width=0.30\textwidth]{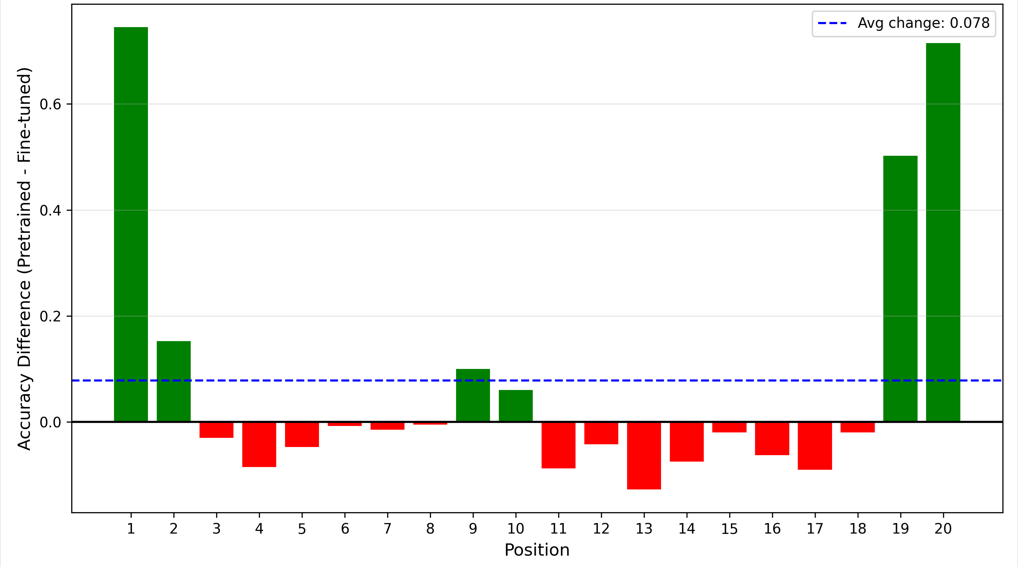} \\
(a) & (b) 
\end{tabular}
\caption{Privacy analysis of BERT model under nucleotide reconstruction attacks:  (a) Nucleotide-wise reconstruction accuracy comparison ; (b) Privacy change after fine-tuning, where positive values (green) indicate improved privacy and negative values (red) indicate decreased privacy}
\label{fig:bert_analysis}
\end{figure*}

The analysis of BERT's vulnerability to nucleotide reconstruction attacks reveals more pronounced privacy implications compared to ALBERT. As shown in Table~\ref{tab:reconstruction_accuracy}, the pre-trained BERT model exhibits extreme vulnerability at sequence endpoints with nearly perfect reconstruction accuracy at positions 1 and 20 (both 1.000), resulting in a higher overall vulnerability (pre-trained: 0.380, fine-tuned: 0.302) compared to random chance (0.25). The fine-tuned model shows substantial privacy improvements, particularly at these endpoint positions, while maintaining a more consistent vulnerability profile across all positions. The reconstruction accuracy reported by Pan et al.'s \cite{pan2020privacy} for BERT pretrained embedding shows an average accuracy of 0.598, significantly higher than our pre-trained BERT model's 0.380 average and our fine-tuned BERT model's 0.302 average.
Fig.~\ref{fig:bert_analysis}(a) breaks down vulnerability by nucleotide type, revealing that fine-tuning significantly improves privacy for  (A), reducing vulnerability from 0.516 to 0.212. Similar improvements are seen for  (C) and  (T), with reductions from 0.343 to 0.315 and 0.430 to 0.347, respectively. Only  (G) shows decreased privacy after fine-tuning, with accuracy increasing from 0.237 to 0.316.
Fig.~\ref{fig:bert_analysis}(b) quantifies these position-specific privacy improvements, demonstrating dramatic enhancements at positions 1, 19, and 20 (large green bars). While positions 3-17 show mixed effects with several slight privacy decreases (red bars), the overall privacy change is substantially positive with an average improvement of 0.078, unlike ALBERT's zero-sum privacy effect. 

\textbf{Key Takeaways for BERT}
\begin{enumerate}
    \item Unlike ALBERT's zero-sum effect, fine-tuning BERT creates a substantial overall privacy improvement with an average change of +0.078 across all positions.
    
    \item BERT exhibits extreme vulnerability at sequence endpoints in its pre-trained state, with perfect reconstruction accuracy at positions 1 and 20, making fine-tuning particularly beneficial for these positions.
    
    \item Fine-tuning BERT improves privacy for three out of four nucleotides (A, C, T), with  showing the most dramatic improvement (from 0.516 to 0.212).
    
    \item The pre-trained BERT model demonstrates higher overall vulnerability (0.380) compared to ALBERT (0.258), suggesting BERT embeddings may inherently preserve more sequence information.
    
    \item For BERT, unlike ALBERT, full fine-tuning appears to be an effective privacy-preserving strategy, though additional nucleotide-specific protections may be needed.
\end{enumerate}

\begin{figure*}[htbp]
\begin{center}
\begin{tabular}{ccc}
\includegraphics[width=0.30\textwidth]{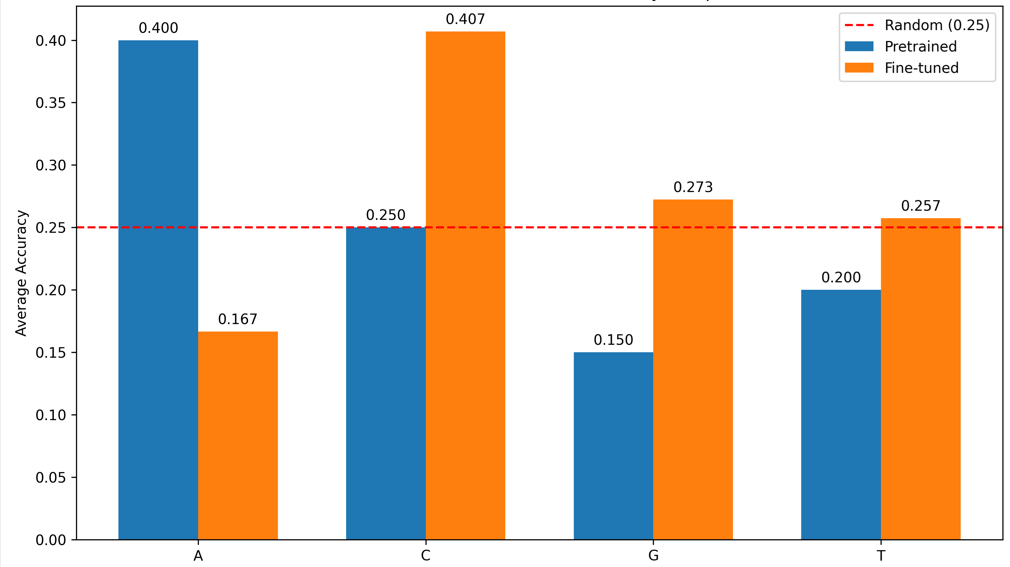} &
\includegraphics[width=0.30\textwidth]{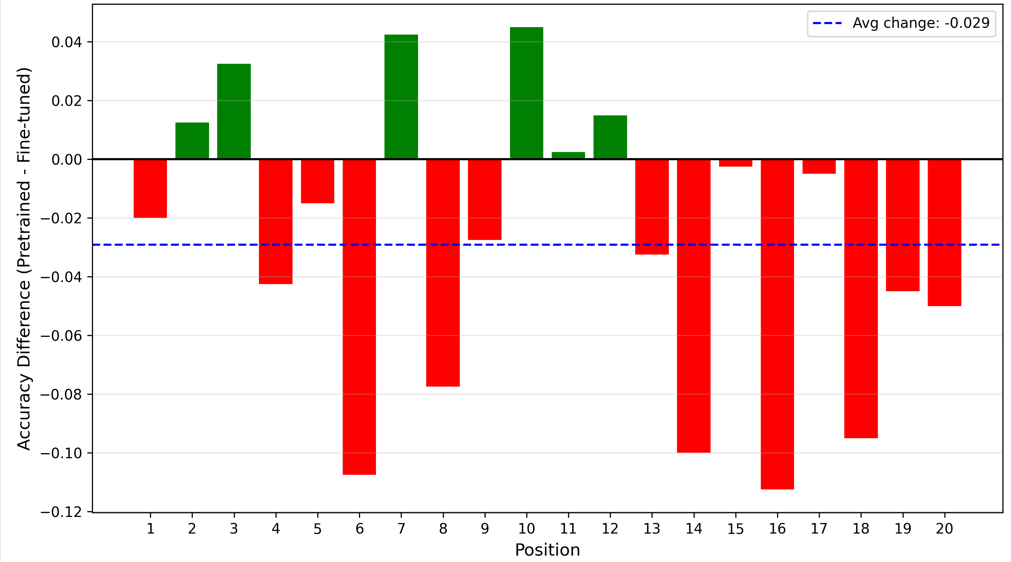} \\
{\fontfamily{ptm}\fontsize{8}{10}\selectfont (a)} & 
{\fontfamily{ptm}\fontsize{8}{10}\selectfont (b)}
\end{tabular}
\end{center}
\caption{Privacy analysis of ERNIE model under nucleotide reconstruction attacks: (a) Nucleotide-wise reconstruction accuracy comparison; (b) Privacy change after fine-tuning where positive values (green) indicate improved privacy and negative values (red) indicate decreased privacy}
\label{fig:ernie_analysis}
\end{figure*}

\subsection{ERNIE Model Privacy Analysis}

The analysis of ERNIE's vulnerability to nucleotide reconstruction attacks reveals privacy characteristics distinctly different from the BERT model. As shown in Table~\ref{tab:reconstruction_accuracy}, the pre-trained ERNIE model operates essentially at a random chance level with an average accuracy of 0.255, suggesting strong inherent privacy protection. Unlike BERT model, ERNIE shows no extreme vulnerability at sequence endpoints, with position values consistently in the 0.190-0.320 range. Notably, fine-tuning increases overall vulnerability to 0.284, representing a privacy deterioration instead of the improvement seen with BERT models.

The reconstruction by Pan et al.'s~\cite{pan2020privacy} for ERNIE pretrained embedding shows an average accuracy of 0.483, which is substantially higher than our pre-trained ERNIE model's 0.255 average and our fine-tuned ERNIE model's 0.284 average.
Fig.~\ref{fig:ernie_analysis}(a) reveals striking nucleotide-specific effects that differ dramatically from the BERT model. Fine-tuning substantially improves privacy for (A), reducing vulnerability from 0.400 to 0.167. However, this comes at the cost of increased vulnerability for the other three nucleotides: (C) increases from 0.250 to 0.407, (G) from 0.150 to 0.273, and (T) from 0.200 to 0.257. This suggests that fine-tuning redistributes vulnerability from A to other nucleotides rather than reducing it.

Fig.~\ref{fig:ernie_analysis}(b) quantifies these position-specific privacy changes, demonstrating that most positions (15 out of 20) experience decreased privacy after fine-tuning (red bars). Positions 6, 14, and 16 show the most substantial privacy deterioration (approximately -0.11). Only five positions (2, 3, 7, 10, 11, 12) show modest privacy improvements (green bars), resulting in an average negative privacy change of -0.029 across all positions.

\textbf{Key Takeaways for ERNIE}
\begin{enumerate}
    \item Pre-trained ERNIE demonstrates strong inherent privacy protection, operating at a random chance level (0.255) without the endpoint vulnerabilities seen in BERT models.
    
    \item Unlike BERT variants, fine-tuning ERNIE decreases overall privacy with an average change of -0.029 across all positions. The privacy impact of fine-tuning is position-dependent, with 75\% of positions showing decreased privacy and only 25\% showing modest improvements.
    
    \item Fine-tuning creates a dramatic tradeoff between nucleotides: significantly improving privacy for (A) while substantially decreasing it for (C), (G), and (T).
    
    \item For ERNIE, full fine-tuning is counterproductive as a privacy-preserving strategy, suggesting that maintaining the pre-trained model or developing nucleotide-specific protection methods would be more effective.
    
    \item ERNIE's privacy degradation during fine-tuning occurs because its knowledge-specialized integration architecture makes it susceptible to this privacy-performance tradeoff, unlike standard transformers. The initially well-balanced parameters that provide strong privacy are disrupted during fine-tuning, as the model optimizes for downstream tasks by redistributing how information is encoded across different nucleotides. 
\end{enumerate}

\subsection{GPT-2 Model Privacy Analysis}
\begin{figure*}[!t]
\centering
\begin{tabular}{ccc}
\includegraphics[width=0.30\textwidth]{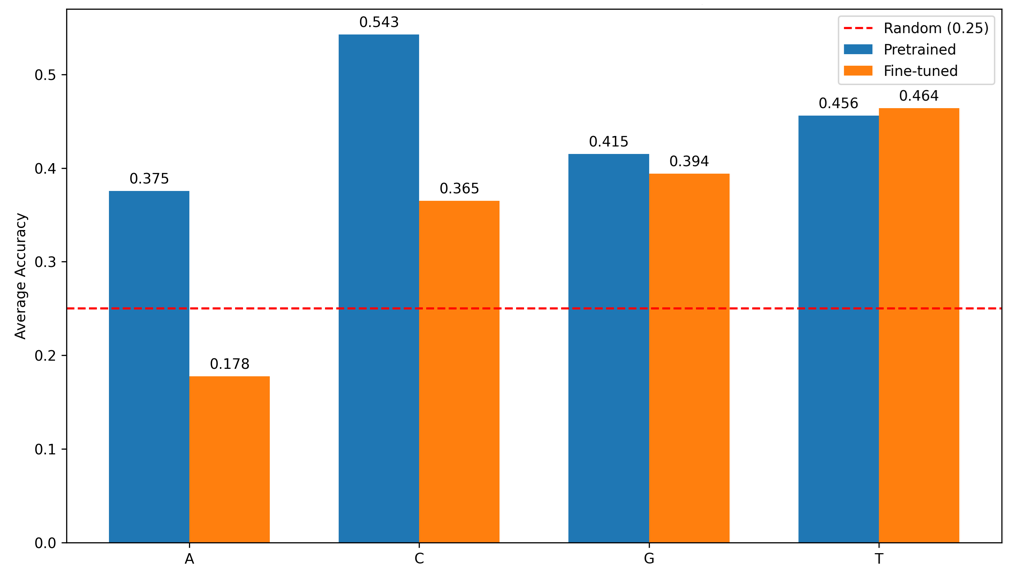} &
\includegraphics[width=0.30\textwidth]{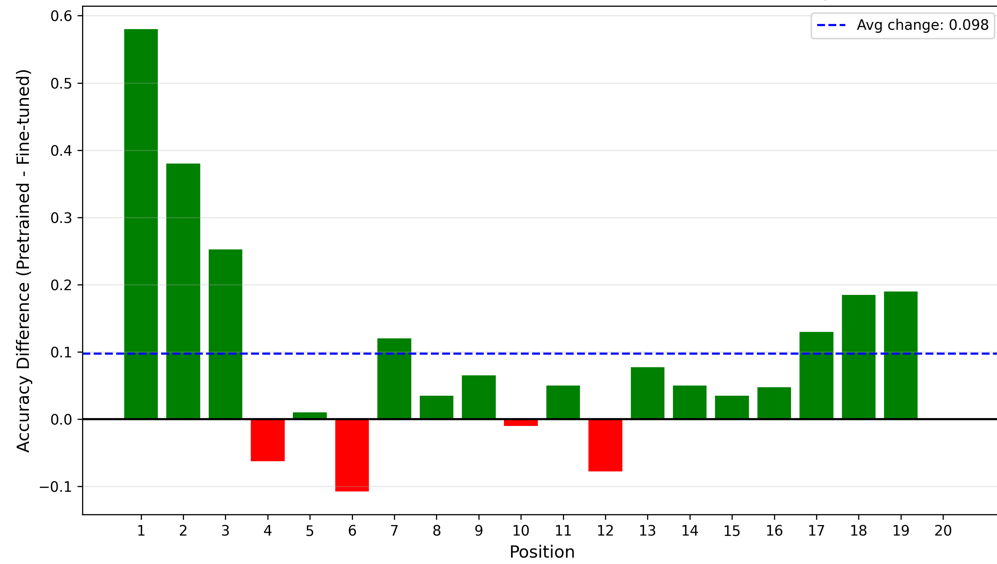} \\
(a) & (b) 
\end{tabular}
\caption{Privacy analysis of GPT-2 model under nucleotide reconstruction attacks: (a) Nucleotide-wise reconstruction accuracy comparison ; (b) Privacy change after fine-tuning, where positive values (green) indicate improved privacy and negative values (red) indicate decreased privacy.}
\label{fig:gpt2_analysis}
\end{figure*}

The analysis of GPT-2's vulnerability to nucleotide reconstruction attacks reveals privacy characteristics that closely resemble the BERT model. As shown in Table~\ref{tab:reconstruction_accuracy}, the pre-trained GPT-2 model exhibits extreme vulnerability at sequence endpoints, particularly positions 1 (0.820), 2 (0.640), 3 (0.520), 19 (0.960), and 20 (1.000), resulting in a high overall vulnerability (pre-trained: 0.450, fine-tuned: 0.352) compared to random chance (0.25). Fine-tuning substantially improves privacy at most positions, especially at the beginning of sequences, but position 20 remains highly vulnerable (0.480) even after fine-tuning. In the comparison by Pan et al.'s \cite{pan2020privacy}, GPT-2 demonstrates an average reconstruction accuracy of 0.624, which is higher than our pre-trained and fine-tuned GPT-2 model's average, with an average of 0.450 and 0.352, respectively. Fig.~\ref{fig:gpt2_analysis}(a) breaks down vulnerability by nucleotide type, revealing that fine-tuning significantly improves privacy for  (A), reducing vulnerability from 0.375 to 0.178, and for  (C), from 0.543 to 0.365.  (G) shows a slight improvement (0.415 to 0.394), while  (T) shows a marginal privacy decrease (0.456 to 0.464). This indicates that fine-tuning primarily benefits A and C nucleotides, with minimal impact on G and T.

Fig.~\ref{fig:gpt2_analysis}(b) quantifies these position-specific privacy changes, demonstrating dramatic improvements at position 1 ($\sim 0.58$
), position 2 ($\sim 0.38$
), and position 3 ($\sim 0.25$
). Most positions (16 out of 20) show privacy improvements after fine-tuning, with only positions 4, 6, 11, and 13 showing minor privacy decreases (red bars). The overall privacy change is substantially positive with an average improvement of 0.098, representing the largest average improvement among all models analyzed.

\textbf{Key Takeaways for GPT-2}
\begin{enumerate}
    \item Despite being an autoregressive model, GPT-2 exhibits extreme endpoint vulnerability similar to the BERT model, suggesting architectural differences significantly impact privacy characteristics.
    
    \item Fine-tuning GPT-2 creates the largest overall privacy improvement (+0.098) of all models analyzed, with 80\% of positions showing enhanced privacy.
    
    \item The most dramatic privacy improvements occur at positions 1-3, while position 20 remains highly vulnerable even after fine-tuning, suggesting persistent information leakage at sequence ends.
    
    \item Fine-tuning substantially improves privacy for (A) and (C), with minimal impact on (G) and  (T), indicating nucleotide-specific effects.
    
    \item For GPT-2, full fine-tuning is an effective privacy-preserving strategy for most positions and nucleotides, but additional position-specific protections may be needed for position 20 and .
\end{enumerate}

\subsection{RoBERTa Model Privacy Analysis}

\begin{figure*}[!t]
\centering
\begin{tabular}{ccc}

\includegraphics[width=0.30\textwidth]{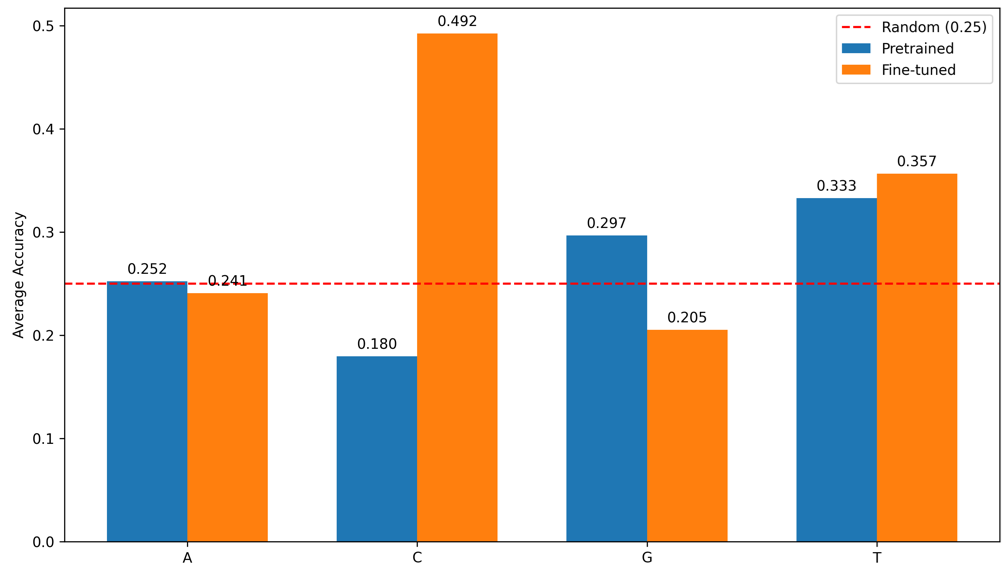} &
\includegraphics[width=0.30\textwidth]{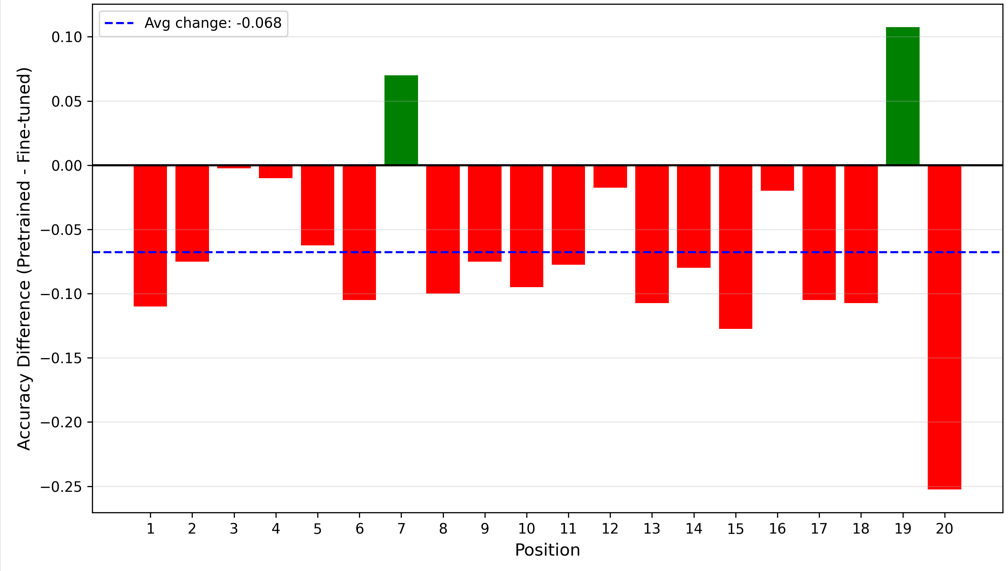} \\
(a) & (b) 
\end{tabular}
\caption{Privacy analysis of RoBERTa model under nucleotide reconstruction attacks:  (a) Nucleotide-wise reconstruction accuracy comparison ; (b) Privacy change after fine-tuning, where positive values (green) indicate improved privacy and negative values (red) indicate decreased privacy}

\label{fig:roberta_analysis}
\end{figure*}

The analysis of RoBERTa's vulnerability to nucleotide reconstruction attacks reveals privacy characteristics distinctly different from both BERT and GPT-2 models. As shown in Table~\ref{tab:reconstruction_accuracy}, the pre-trained RoBERTa model operates close to random chance level with an average accuracy of 0.264, suggesting strong inherent privacy protection. Unlike BERT variants, RoBERTa shows no extreme vulnerability at sequence endpoints, with most positions showing accuracy between 0.190 and 0.340, except for position 19, which reaches 0.450. However, fine-tuning substantially increases overall vulnerability to 0.331, representing a significant privacy deterioration of -0.067. In the comparison by Pan et al.'s \cite{pan2020privacy}, RoBERTa shows a reconstruction accuracy of 0.380, which is higher than our pre-trained RoBERTa model's 0.264 average but comparable to our fine-tuned model's 0.331 average.
Fig.~\ref{fig:roberta_analysis}(a) reveals dramatic nucleotide-specific effects. Fine-tuning slightly improves privacy for (A), reducing vulnerability from 0.252 to 0.241, and substantially improves privacy for (G), from 0.297 to 0.205. However, this comes at the cost of dramatically decreased privacy for (C), which increases from 0.180 to 0.492 (the highest nucleotide-specific vulnerability observed in any model), and slightly decreased privacy for (T), from 0.333 to 0.357.
Fig.~\ref{fig:roberta_analysis}(b) quantifies these position-specific privacy changes, demonstrating that 90\% of positions (18 out of 20) experience decreased privacy after fine-tuning (red bars). Position 20 shows the most dramatic privacy deterioration (approximately -0.25), while only positions 7 and 19 show modest privacy improvements (green bars). This results in an average negative privacy change of -0.068 across all positions, representing the largest privacy degradation among all models analyzed.

\begin{figure*}[!t]
\centering
\begin{tabular}{ccc}

\includegraphics[width=0.30\textwidth]{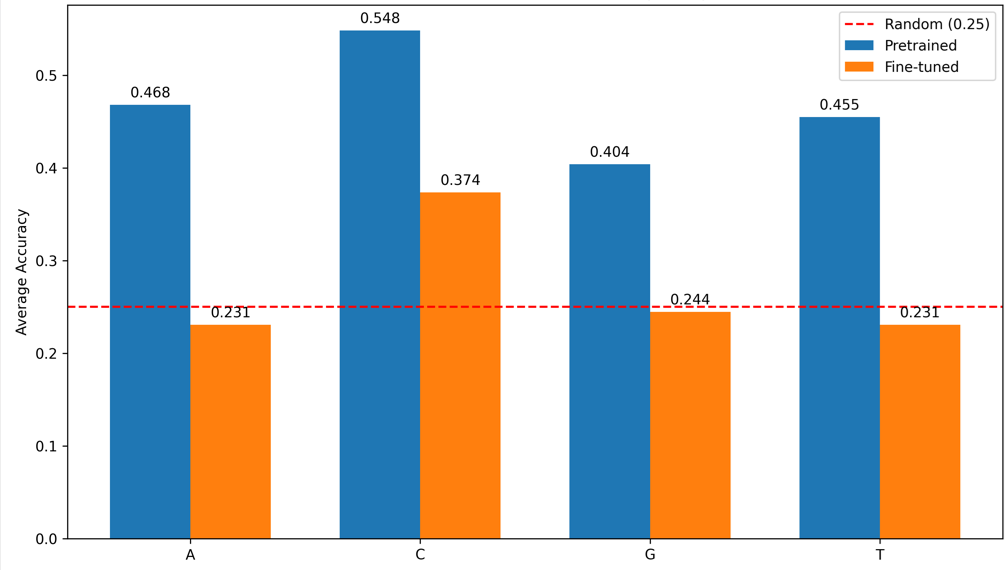} &
\includegraphics[width=0.30\textwidth]{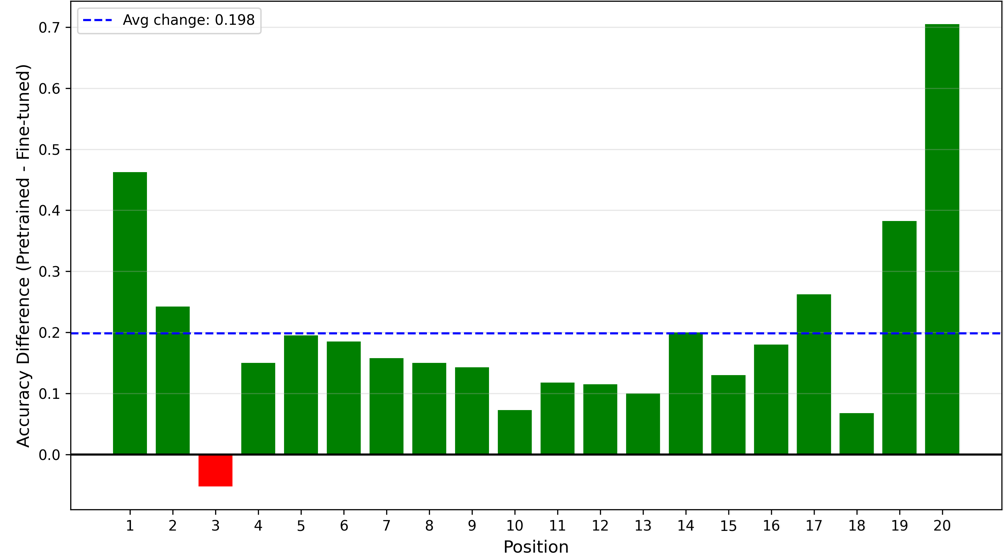 } \\
(a) & (b) 
\end{tabular}
\caption{Privacy analysis of XLNet model under nucleotide reconstruction attacks: (a) Nucleotide-wise reconstruction accuracy comparison showing significant privacy improvements for all nucleotides after fine-tuning; (b) Privacy change after fine-tuning where positive values (green) indicate improved privacy and negative values (red) indicate decreased privacy.}
\label{fig:xlnet_analysis}
\end{figure*}

\textbf{Key Takeaways for RoBERTa}
\begin{enumerate}
    \item Pre-trained RoBERTa demonstrates strong inherent privacy protection, operating close to random chance level (0.264) without the endpoint vulnerabilities seen in BERT model.
    
    \item Unlike most models analyzed, fine-tuning RoBERTa substantially decreases overall privacy with an average change of -0.068 across all positions. The privacy impact of fine-tuning is overwhelmingly negative across positions, with 90\% of positions showing decreased privacy and only 10\% showing modest improvements.
    
    \item Fine-tuning creates a dramatic tradeoff between nucleotides: improving privacy for (A) and (G) while substantially decreasing it for (C) and slightly decreasing it for (T). RoBERTa's unique pre-training approach creates initially privacy-preserving representations that are particularly vulnerable to disruption during fine-tuning, with cytosine (C) nucleotides showing the most dramatic vulnerability increase.
    
    \item In RoBERTa, privacy degradation observed during fine-tuning arises from its pre-training methodology, which integrates dynamic masking with large batches. These strategies build strong initial privacy protection, yet that protection can diminish when the model is optimized for particular tasks. The finding underscores that models starting with high privacy protection may become more vulnerable as they are fine-tuned.
    
    \item For RoBERTa, full fine-tuning is counterproductive as a privacy-preserving strategy, suggesting that maintaining the pre-trained model or developing position-specific and nucleotide-specific protection methods would be more effective.
\end{enumerate}

\subsection{XLNet Model Privacy Analysis}

The analysis of XLNet's vulnerability to nucleotide reconstruction attacks reveals privacy characteristics that mirror BERT model but with even more pronounced effects. As shown in Table~\ref{tab:reconstruction_accuracy}, the pre-trained XLNet model exhibits extreme vulnerability at sequence endpoints, particularly positions 1 (0.720), 17 (0.570), 19 (0.660), and 20 (0.970), resulting in a high overall vulnerability (pre-trained: 0.471, fine-tuned: 0.273) compared to random chance (0.25). Fine-tuning dramatically improves privacy across all positions, transforming the pronounced U-shaped vulnerability pattern into a nearly flat profile hovering just above random chance. In the comparison by Pan et al.'s \cite{pan2020privacy}, XLNet demonstrates a reconstruction accuracy of 0.624, which is substantially higher than our pre-trained XLNet model's 0.471 average and significantly higher than our fine-tuned model's 0.273 average.
Fig.~\ref{fig:xlnet_analysis}(a) breaks down vulnerability by nucleotide type, revealing that fine-tuning significantly improves privacy for all four nucleotides. The most substantial improvements occur for  (A), reducing vulnerability from 0.468 to 0.231, and  (T), from 0.455 to 0.231.  (C) and  (G) also show significant improvements, from 0.548 to 0.374 and from 0.404 to 0.244, respectively. This indicates that fine-tuning benefits all nucleotides, with A and T approaching random chance levels after fine-tuning.
Fig.~\ref{fig:xlnet_analysis}(b) quantifies these position-specific privacy changes, demonstrating remarkable improvements at nearly all positions (19 out of 20), with only position 3 showing a minor privacy decrease. The most dramatic enhancements occur at position 1 (approximately 0.46) and position 20 (approximately 0.70). The overall privacy change is substantially positive with an average improvement of 0.198, representing the most significant average improvement among all models analyzed.

\textbf{Key Takeaways for XLNet}
\begin{enumerate}
    \item Fine-tuning XLNet creates the most substantial overall privacy improvement (+0.198) of all models analyzed.
    
    \item Pre-trained XLNet exhibits extreme endpoint vulnerability similar to the BERT model, but with even higher reconstruction accuracy at position 20 (nearly 1.0).
    
    \item Fine-tuning improves privacy for all four nucleotides, with  (A) and  (T) approaching random chance levels after fine-tuning.
    
    \item The privacy impact of fine-tuning is overwhelmingly positive across positions, with 95\% of positions showing improved privacy.
    
    \item For XLNet, full fine-tuning is extremely effective as a privacy-preserving strategy, transforming a highly vulnerable model into one that operates close to random chance levels for most positions and nucleotides.
\end{enumerate}

\section{Conclusion and Future Work}
In this study, we investigated reconstruction attacks on language models' embeddings processing genomic data, extending Pan et al.'s methodology \cite{pan2020privacy} to task-specialized models.  We reveal essential insights into how domain adaptation affects privacy vulnerabilities in genomic sequence analysis. Our position-specific vulnerability analysis demonstrated striking differences between model architectures, with the BERT model exhibiting pronounced U-shaped vulnerability patterns with extreme susceptibility at sequence endpoints, while models like ERNIE showed more uniform vulnerability distributions. Our nucleotide-specific analysis revealed that fine-tuning affects different nucleotides asymmetrically, often creating privacy tradeoffs where protection for one nucleotide comes at the expense of others.

Our findings establish a clear model-dependent relationship between fine-tuning and privacy protection. While fine-tuning dramatically improves privacy for specific architectures (XLNet: +19.8\%, GPT-2: +9.8\%, BERT: +7.8\%), it significantly degrades privacy for others (RoBERTa: -6.8\%, ERNIE: -2.9\%). We recommend position-specific and nucleotide-specific protection strategies tailored to each model's unique vulnerability profile rather than applying uniform privacy approaches across all architectures.

Future work should focus on developing targeted protection mechanisms for vulnerable nucleotide representations while preserving model utility and evaluating how parameter-efficient fine-tuning techniques (e.g., LoRa and QLoRa) affect privacy-utility trade-offs in nucleotide reconstruction.

\bibliographystyle{IEEEtran}
\bibliography{ref}
\end{document}